\documentclass[sigconf]{acmart}
\renewcommand\footnotetextcopyrightpermission[1]{} 
\setcopyright{none}

\usepackage{amsmath}

\usepackage{algorithmic}
\usepackage{graphicx}
\usepackage{textcomp}
\usepackage{float}

\usepackage[nolist,nohyperlinks]{acronym}

\usepackage{xspace}
\newcommand{\latex}{\LaTeX\xspace}

\usepackage{xcolor}
\usepackage{siunitx}
\DeclareSIUnit\pixel{px}

\AtBeginDocument{%
  \providecommand\BibTeX{{%
    \normalfont B\kern-0.5em{\scshape i\kern-0.25em b}\kern-0.8em\TeX}}}

\acmConference[]{}{}{}
\acmBooktitle{}

\begin{document}

\title{Unsupervised Training Data Generation of Handwritten Formulas using Generative Adversarial Networks with Self-Attention}


\author{Matthias Springstein}
\email{matthias.springstein@tib.eu}
\orcid{1234-5678-9012}
\affiliation{%
  \institution{TIB -- Leibniz Information Centre for Science and Technology}
  \city{Hannover}
  \country{Germany}
}

\author{Eric Müller-Budack}
\email{eric.mueller@tib.eu}
\orcid{1234-5678-9012}
\affiliation{%
  \institution{TIB -- Leibniz Information Centre for Science and Technology}
  \city{Hannover}
  \country{Germany}
}

\author{Ralph Ewerth}
\email{ralph.ewerth@tib.eu}
\orcid{1234-5678-9012}
\affiliation{%
  \institution{TIB -- Leibniz Information Centre for Science and Technology}
  \institution{L3S Research Center, Leibniz University Hannover}
  \city{Hannover}
  \country{Germany}
}








\begin{abstract}

The recognition of handwritten mathematical expressions in images and video frames is a difficult and unsolved problem yet. Deep convectional neural networks are basically a promising approach, but typically require a large amount of labeled training data. However, such a large training dataset does not exist for the task of handwritten formula recognition. In this paper, we introduce a system that creates a large set of synthesized training examples of mathematical expressions which are derived from \latex documents. For this purpose, we propose a novel attention-based generative adversarial network to translate rendered equations to handwritten formulas. 
The datasets generated by this approach contain hundreds of thousands of formulas, making it ideal for pretraining or the design of more complex models. 
We evaluate our synthesized dataset and the recognition approach on the CROHME 2014 benchmark dataset. Experimental results demonstrate the feasibility of the approach.

\end{abstract}






\maketitle

\section{Introduction}\label{sec:intro}
In recent years, deep neural networks have been successfully applied in many domains of computer vision and natural language processing. Most of the progress is enabled by the fact that more complex models can be trained with a larger number of layers and consequently, more parameters. State-of-the-art models like ResNet~\cite{he2016deep} or DenseNet~\cite{huang2017densely} for the 2012 ILSVRC~(ImageNet Large Scale Visual Recognition Competition~\cite{ILSVRC15}) challenge include several million parameters. In order to prevent these models from overfitting, various techniques have been introduced in recent years, such as weight decay, batch normalization or different data augmentation techniques. 
Another way to avoid overfitting is to increase the number of samples or, if that is not possible, to pretrain the model with a larger data set that has been created for a similar problem. 
Sun et al.~\cite{sun2017revisiting} show that the collection of more examples helps to improve the performance of deep neural networks.

\begin{figure}
  \centering
\includegraphics[width=1.0\linewidth]{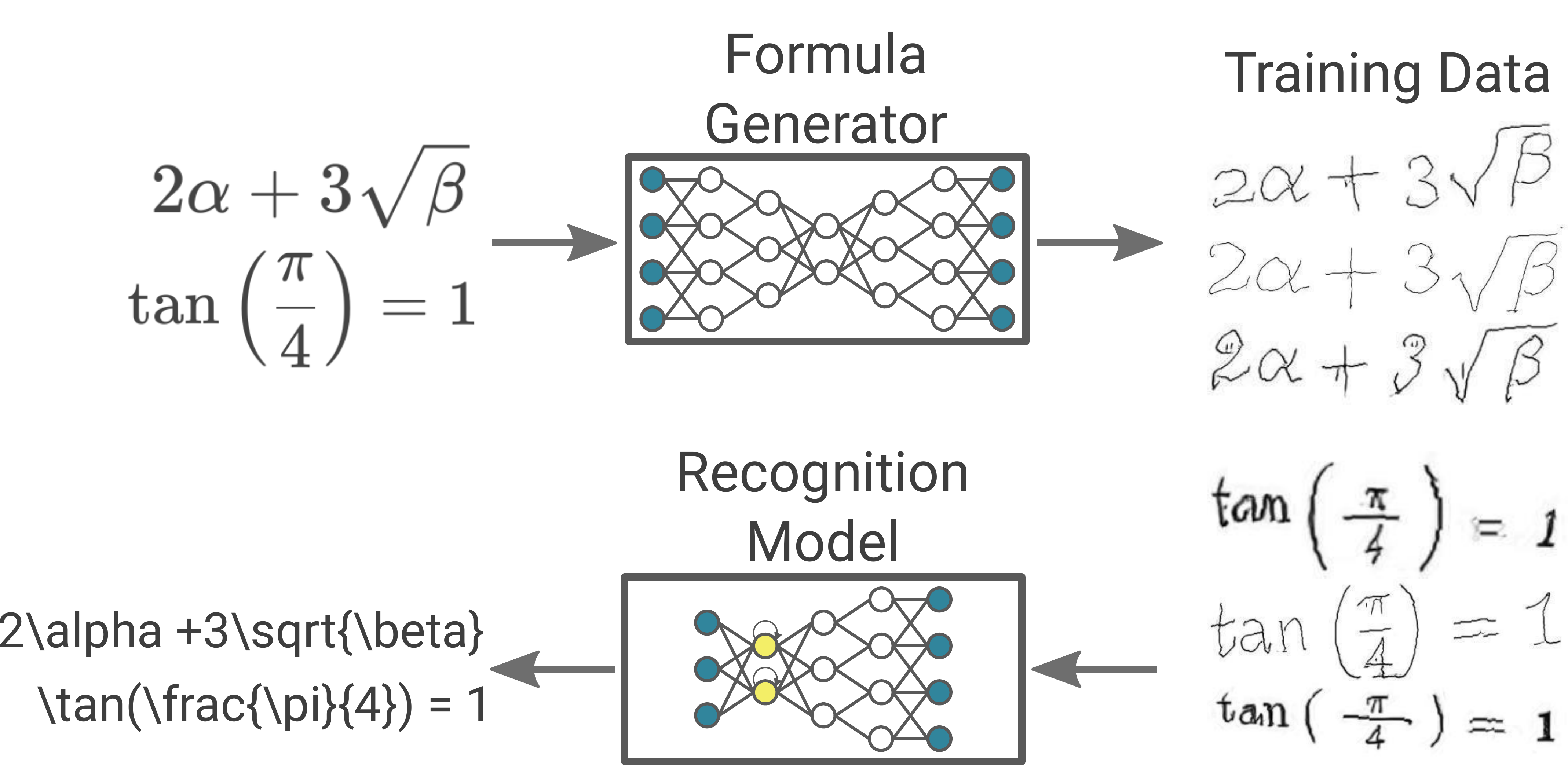}
\caption{Illustration of the two-step approach to increase the amount of example data to train formula recognition models. Top: training of a GAN to synthesize handwritten formulas. Bottom: use of the generated formulas for the training of the actual formula recognition model.}
\label{fig:paper_idea}
\end{figure}

For the task of formula recognition in images, it is easy to collect machine generated examples created using the \latex suite. On the one hand, scientists from different disciplines use this system to create and to share their research results with their colleagues, resulting in a large collection of machine generated formulas. On the other hand, it is very difficult to find comparable collections of handwritten equations and symbols. The largest available datasets for handwritten equations contain only several thousand examples. Collecting additional examples is a time-consuming and expensive task, since it requires pen-based input systems and a certain degree of diversity of participants. The problem gets even more difficult when the length of the target sequence is increased or more complicated symbols are added.

In this paper, we introduce a new attention-based generative adversarial network~(GAN) that easily translates rendered \latex equations to handwritten ones. To achieve this, we combine a current self-attention GAN~\cite{zhang2018self} implementation with an recurrent neural networks~(RNN) based task classifier to enable pixel-to-pixel transformations. The final system is able to generate a large set of examples for the training of handwritten formula recognition models and thus, improves the accuracy of such systems. The self-attention module in our generator allows the model to learn from regional features that are (far) distant from one another. This helps our model to learn symbols that occur in pairs, like brackets or symbols that occupy a larger area, like \emph{\textbackslash sqrt} and \emph{\textbackslash frac}. Experimental results on the benchmark dataset CROHME demonstrate the feasibility of the approach. The code and the synthesized datasets are available at \url{https://github.com/TIBHannover/formula_gan}. 
Due to the comparably huge size of the generated dataset and the variation of the shown formulas, the generated dataset is suitable for the pretraining of more complex models.

The remainder of the paper is organized as follows. Section \ref{sec:related} reviews related work on formula recognition and GANs. Section \ref{sec:model} gives an overview of the proposed models, while the used datasets and pre-processing steps are described in Section \ref{sec:data}. The experimental setup and results are presented in Section \ref{sec:exp}. Section \ref{sec:conclusion} draws a brief conclusion and gives an outlook on future work.

\section{Related Work}\label{sec:related}
This section reviews related work on mathematical formula recognition~(section~\ref{sec:rw_formularec}) and Generative Adversarial Networks~(Section~\ref{sec:rw_gan}), which have been widely applied to generate synthetic images in recent years. 


\subsection{Formula Recognition}\label{sec:rw_formularec}

The task of formula recognition can be divided into two groups: online and offline recognition. For online recognition of formulas, not only an image is available for recognition, but also temporal information represented by a sequence of strokes. This task is more relevant for office software or tablet users who draw equations on the device. We do not consider this use case in this paper.

Current approaches for offline formula recognition rely on deep neural networks to recognize text or symbols in images. These approaches use an encoder-decoder structure with attention techniques to link between the two models. The task of the encoder part is to extract useful features from the input image and the decoder tries to predict the corresponding sequence of tokens based on the extracted features. One of the first approaches of this kind was proposed by Deng et al.~\cite{deng2017image,deng2016wygiswys}. They extracted \num{100000} formulas from scientific publications and used the found \latex formulas to train their models. Deng et al.~\cite{deng2017attention} extend this general approach by adding two additional recurring neural networks that learn horizontal and vertical embeddings of the input image. Zhang et al.~\cite{zhang2019improved} extract image information with an CNN on three different scales and use two attention layers in order to increase the accuracy for small symbols. Recently, Wang and Liu~\cite{DBLP:journals/corr/abs-1908-11415} proposed a stacked LSTM~\cite{hochreiter1997long} decoder fed by a CNN network with position embedding with features

For the recognition of handwritten mathematical formulas, the number of annotated data is significantly lower and therefore strong augmentation and regularization techniques have to be used to prevent overfitting. Zhang et al.~\cite{zhang2017watch} used a convolutional neural network for their approach to extract the features jointly with a GRU~(Gated Recurrent Unit) and an attention-based decoder. In subsequent work, Zhang et al.~\cite{zhang2018multi} relied on a DenseNet architecture~\cite{huang2017densely} and extracted features from different scales to improve accuracy. Other approaches aim at recognizing the individual symbols in the formula. For example, Davila and Zanibbi~\cite{davila2018visual} suggested to detect individual symbols and use line-of-sight graphs as a retrieval approach. Le et al.~\cite{le2019pattern} use a set of local and global transformations for the individual characters and the entire equation to extend the training dataset and improve the generalization of the model. Wu et al.~ \cite{wu2020handwritten} fed rendered images into the classification network in parallel to handwritten images to find a common embedding for both domains. One main difference of our method is that we aim to create a new dataset that should be suitable for any architecture. 

\subsection{Generative Adversarial Networks}\label{sec:rw_gan}

Generative adversarial networks have achieved impressive results in various computer vision tasks, such as image generation and image-to-image translation. Based on Goodfellow et al.'s architecture~\cite{goodfellow2014generative} several extensions were made to allow such networks to respond to conditions~\cite{mirza2014conditional} or to generate images with higher resolution~\cite{miyato2018spectral,brock2018large}, so that it is now possible to create realistic images for the ILSVRC 2012 concepts~\cite{ILSVRC15}.

Recently, Miyato et al.~\cite{miyato2018spectral} introduced a new method to regularize the spectral norm of the weight matrix in the discriminator to stabilize the training behavior of the GAN and allow the model to generalize well on the ILSVRC 2012 concepts~\cite{ILSVRC15}. One of the drawbacks of this method is that the synthesized images look right, but the alignment of the single components does not look realistically. To solve this issue, Zhang et al.~\cite{zhang2018self} proposed a self-attention layer that enables the generator and the discriminator to rely on non-local features and to learn a spatial routing of features. Brock et al.~\cite{brock2018large} investigated the behavior of conditional GANs by increasing the number of parameters and the number of examples per iteration and were able to show significant performance improvements. In addition, the authors investigate some techniques to reduce instabilities in the training of large-scale GANs.

Another way to create handwritten text is to generate the strokes of the individual symbols instead of calculating the image pixel by pixel. This makes it possible to transform the task into a sequence to sequence translation problem. The first approaches \cite{graves2013generating,carter2016experiments,kumar2018synthesizing} towards this direction use RNN and Mixture Density Networks for the prediction of the individual stroke coordinates. A limitation of this approach is that the pure stroke information is very idealized and cannot include distortion.

\section{A GAN Model for Formula Synthesis}\label{sec:model}

\begin{figure*}
  \centering
\includegraphics[width=.85\textwidth]{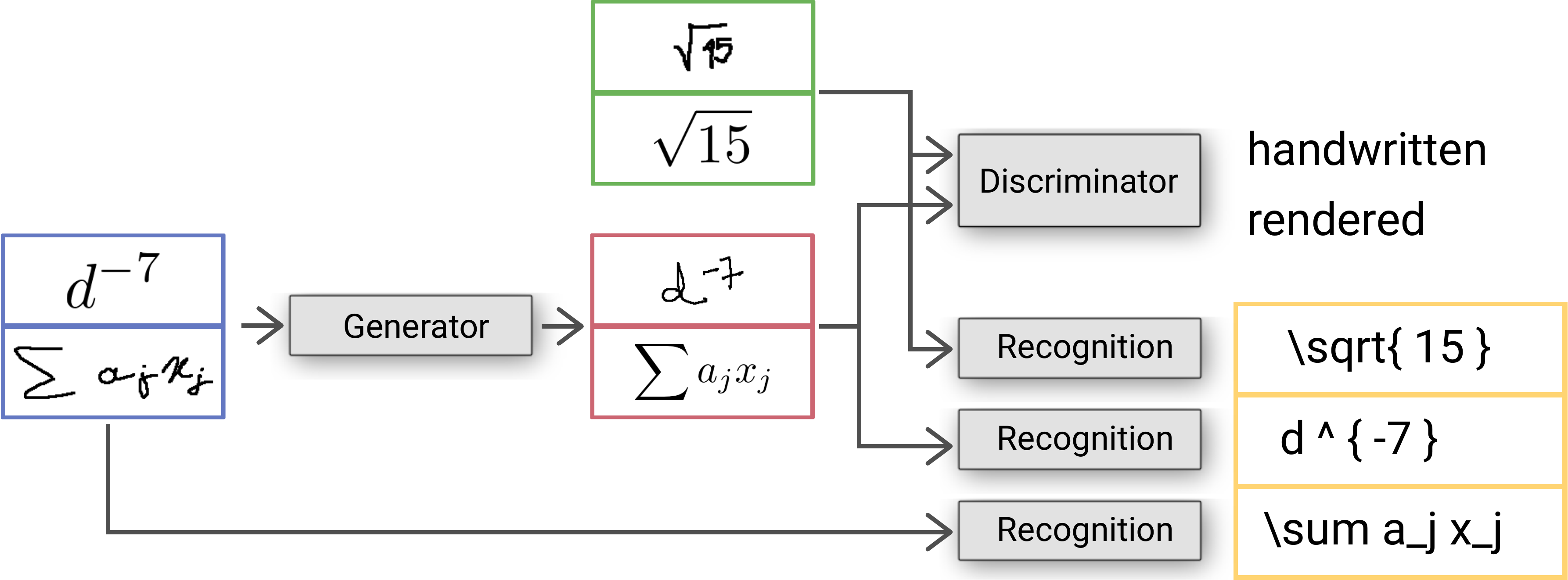}
\caption{Structure of the GAN: The blue equations are samples from the source distribution and are fed to the generator so that the generated red equation looks similar to the green target distribution. All distributions are used to train the corresponding recognition task models.}
\label{fig:gan_model}
\end{figure*}

In this section, we present the proposed framework for translating an image of a formula into a handwritten representation. The main task of the model is to preserve the components of the respective \latex sequence during the translation from one domain to another. To achieve this goal, we extend Bousmalis et al.'s approach~\cite{bousmalis2016unsupervised} and combine a GAN with a recognition model, which checks if every character present in one domain still exists in the other one. The recognition model punishes the generator during the training of the GAN if symbols are falsified, replaced, or omitted. An overview of the proposed framework is presented in Figure~\ref{fig:gan_model}.

%

\subsection{Image-to-Image Model}\label{sec:translate_model}

The task of the image-to-image model is to translate an image of arbitrary size from the source to the target domain. Our proposed approach builds upon Brock et al.'s model ~\cite{brock2018large}. In contrast to the original model, the proposed architecture is extended with an encoder structure similar to the discriminator, but without the self-attention layer. On the one hand this can save resources if only one attention layer is used in the encoder, on the other hand, Brock et al.~\cite{brock2018large} show that several attention modules could not further improve the performance.


\begin{figure*}
  \centering
\begin{tabular}{ccc}
\includegraphics[height=8.5cm]{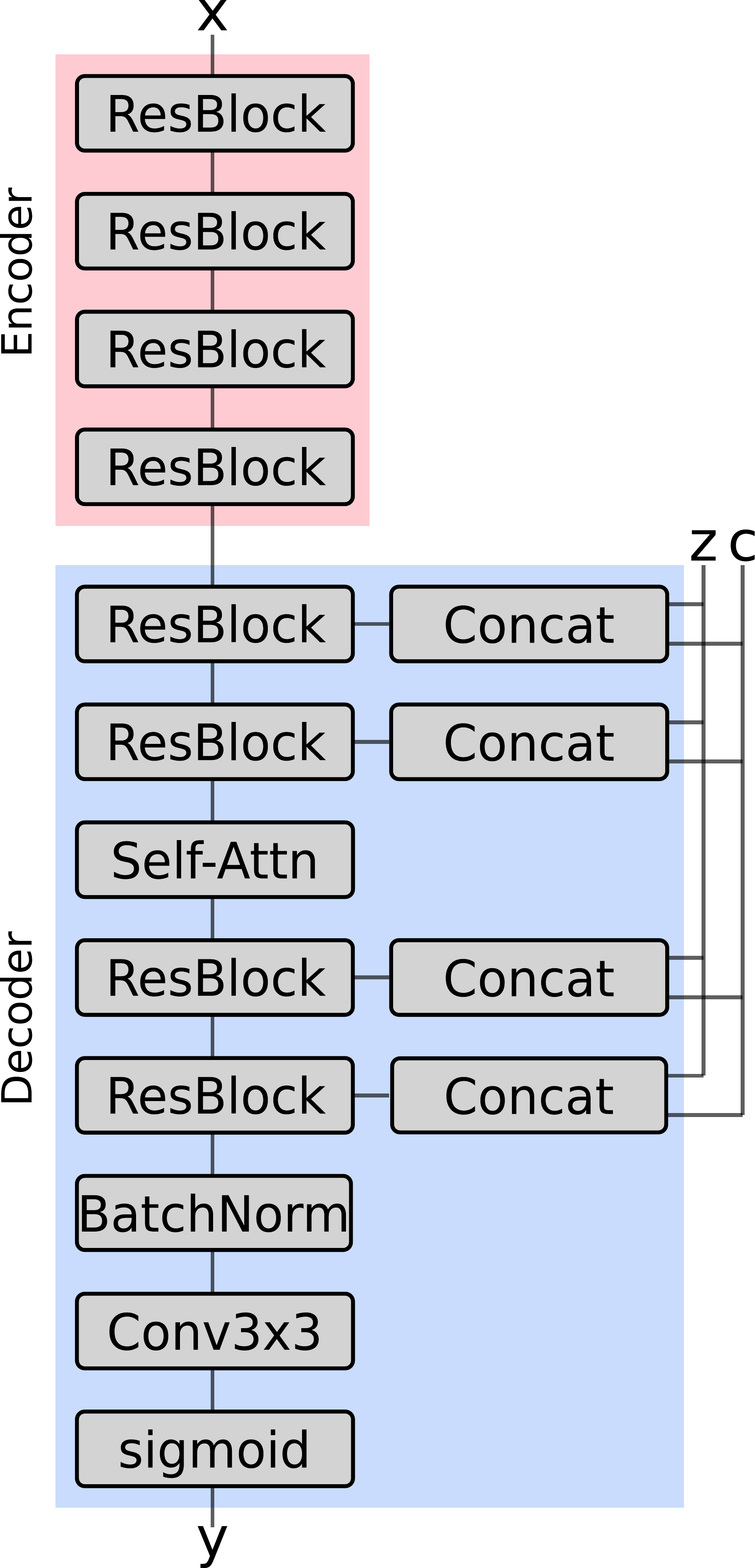}&
\includegraphics[height=8.5cm]{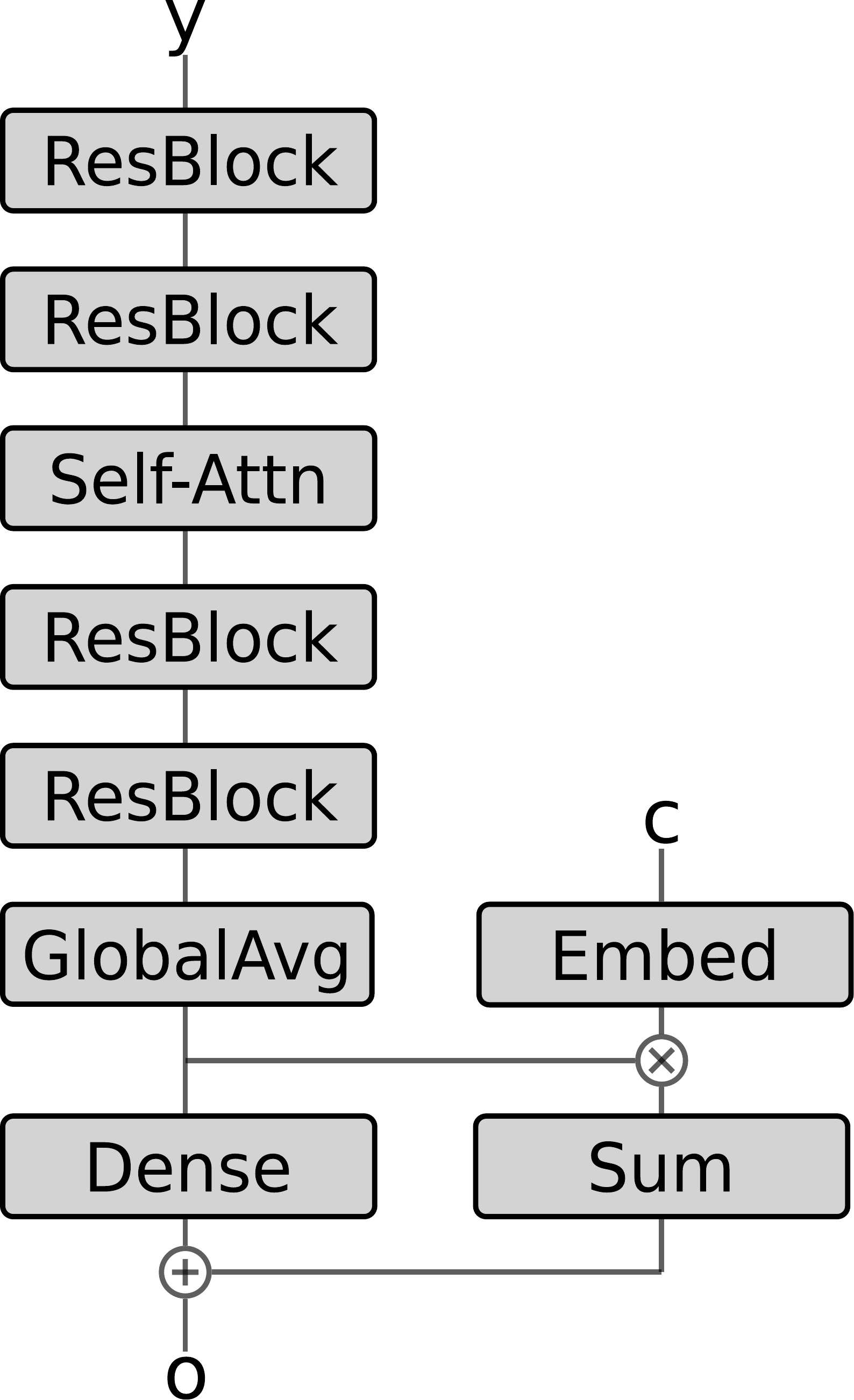}&
\includegraphics[height=8.5cm]{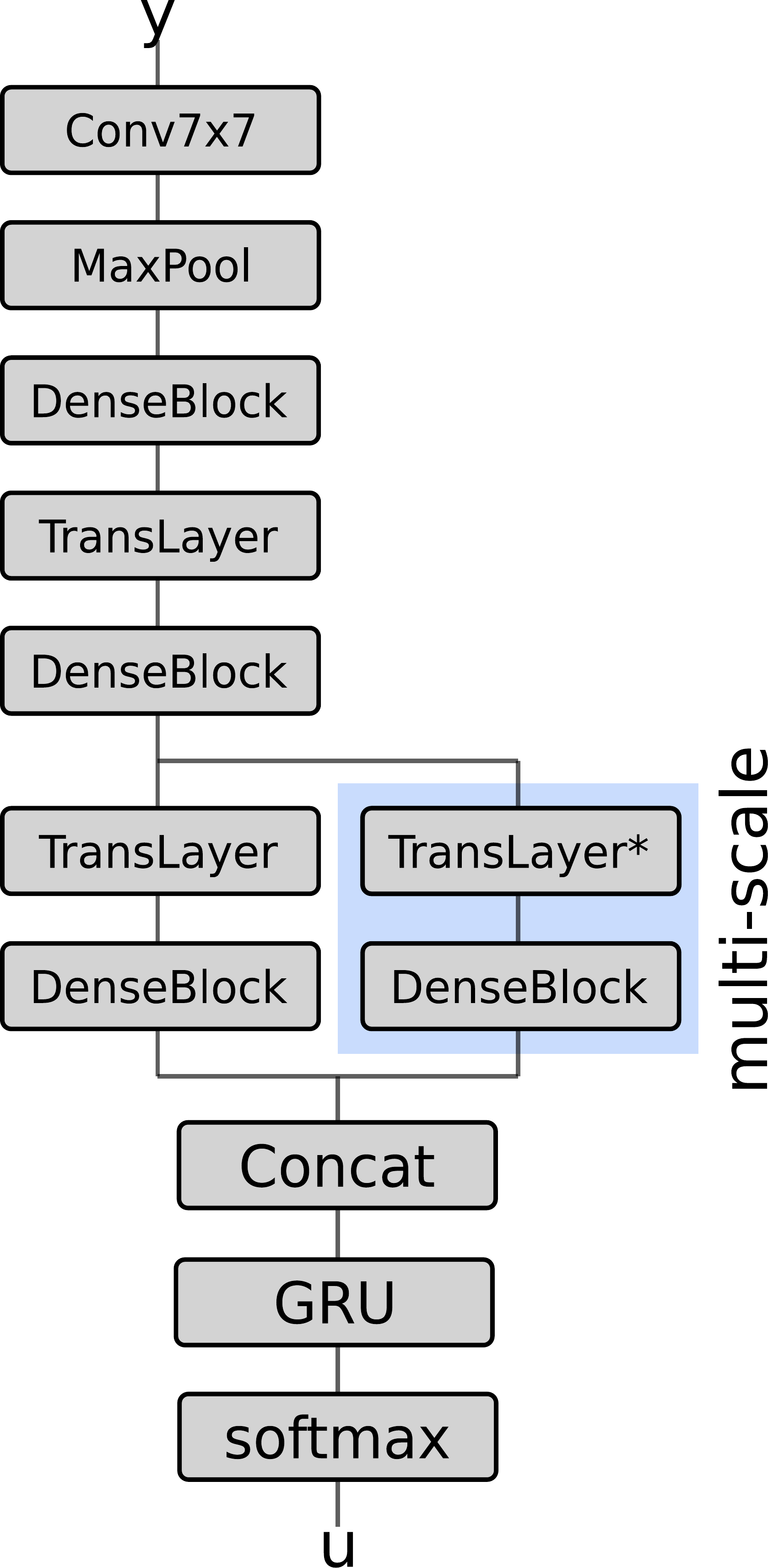}\\
(a)&(b)&(c)
\end{tabular}
\caption{(a) Encoder and decoder part of the generator model. (b) Discriminator model with projection layer~\cite{miyato2018cgans}. (c) Task model with Gated Recurrent Unit (GRU) cell to predict the \latex sequence.}
\label{fig:part}
\end{figure*}


In addition to the actual task of the generator to convert rendered formulas into handwritten ones, there is also the possibility to learn the opposite domain transfer. Liu et al.~\cite{liu2017unsupervised} show that a model that trained using several different transformations can achieve better performance. Therefore, we also want to evaluate this training pattern for the synthesis of handwritten formulas. In our opinion, this approach has several advantages. First, the generator is enforced to find a common embedding for both classes, for the handwritten and the rendered \latex equations. Second, the discriminator only distinguishes between both domains to learn the class assignments of both distributions. Therefore, we prevent the generator from not transferring the domain of the input image at all in order to fool the discriminator.


The encoder consists of four ResBlock modules. Each block is made up of the sequence of \textit{Conv3x3} (convolution), \textit{ReLU} (rectified linear unit), \textit{Downsample}, and \textit{Conv3x3} layers, followed by the decoder which consists of four ResBlock modules with an additional self-attention layer. In contrast to the encoder, an upsampling is performed in each block so that the generated image has the same resolution as the input image. Additionally, we add a BatchNorm layer before each ReLU layer in the decoder, and use a linear projection from the random value $z$ and the condition $c$ to calculate the BatchNorm gains and biases. The end of the model forms a normal BatchNorm without conditioning and Conv3x3 layer and due to the nature of the task, a sigmoid activation function.
The architecture of the entire generator is illustrated in Figure \ref{fig:part} (a).

The discriminator is symmetrically constructed to the decoder of the generator and also uses four residual blocks with a self-attention layer. A projection layer as proposed by Miyato and Koyama~\cite{miyato2018cgans} is used to determine the class (handwritten or machine-generated). The entire discriminator is illustrated in Figure \ref{fig:part} (b).

We use the hinge version of the adversarial loss $L_D$ and $L_G$~\cite{zhang2018self} to optimize the generator $G$ (Eq. \ref{eq:l_d}) and discriminator $D$ (Eq. \ref{eq:l_g}).

\begin{align}
  L_D =& -\mathbb{E}_{(y,c)\sim p_{data}} \left[\min\left(0, -1 +D\left(y,c\right)\right)\right]\label{eq:l_d} \\ \nonumber
  &-\mathbb{E}_{z\sim p_z,c\sim p_{data}}\left[\min\left(0, -1 -D\left(G\left(x, z\right),c\right)\right)\right]\\
  L_G =& -\mathbb{E}_{z\sim p_z,c\sim p_{data}}D\left(G\left(x, z\right)\right)\label{eq:l_g}
\end{align}

\subsection{Self-Attention Module}\label{sec:self_attention}

The self-attention layers allow the model to learn an internal routing of features based on the current content of the previous layers. This is achieved because the output of each position corresponds to a weighted sum of the features from the preceding layer. In our proposed models we use the attention mechanism as described by Zhang et al.~\cite{zhang2018self}.

\begin{align}
  s_{ij} &= f\left(x_i\right)^T g\left(x_j\right)\\
  \beta_{j,i}&= \frac{\exp\left(s_{ij}\right)}{\sum_{i=1}^N\exp\left(s_{ij}\right)}\\
  o_j &= \sum_{i=1}^N \beta_{j,i} h\left(x_i\right)
\end{align}

For a given input $x$ from a previous layer, three feature maps  $f\left(x_i\right)$, $g\left(x_j\right)$ and $h\left(x_i\right)$ are generated with the help of a 1x1 convolution layer function. Here, the feature maps of $f$ and $g$ are used to calculate the importance of a region $i$ for another region $j$ in the matrix $s$.  Finally, a softmax function is used to compute the attention map $\beta$ and thus the output of the attention module $o_j$. 


\begin{align}
  y_i &= \gamma o_j+x_i\label{eq:self_result}
\end{align}

The result of the self-attention layer (Eq. \ref{eq:self_result}) is the weighted sum of the input of the layer $x_i$ and the result of the operation $o_j$. The parameter $\gamma$ is initialized as zero, so that the network only relies on local features in the first training steps.

\subsection{Task Model}\label{sec:task_model}

The recognition model takes an image of a formula as input and predicts a corresponding \latex token sequence. Two different models are used for the proposed approach. The task for the first model is to ensure that, during the training of the GAN, the results of the generator still contain the same token sequence and that no symbols are missing or distorted. Since the second model only trains the recognition of formulas and no GAN needs to be trained, we extend the number of parameters of the model in order to achieve the highest possible performance. Both models are based on the work of Zhang et al.~\cite{zhang2018multi} with a DenseNet encoder architecture and a decoder based on a GRU with Bahdanau attention \cite{bahdanau2014neural}. 
The structure of the network is illustrated in Figure \ref{fig:part} (c).

The optimization objective of both models is to reduce the softmax cross-entropy function, that is the task loss $L_T$, as defined by equation \ref{eq:task_loss}. In this formula, $u$ stands for the predicted symbol at time $t$ and $\hat{u}$ represents the corresponding ground truth label.

\begin{align}
L_T &= -\sum_{t=1}^T \hat{u}_{t}\log(u_{t})\label{eq:task_loss}
\end{align} 

The final loss for both the generator and the discriminator ($L_{GT}$ and $L_{DT}$) are defined as the combination of the non-adversarial task loss~$L_T$ and the corresponding adversarial losses:

\begin{align}
  L_{DT} = L_D + \lambda L_T\\
  L_{GT} = L_G + \lambda L_T
\end{align}

During GAN training a smaller version of the original recognition model was used (omitting the blue box in Figure~\ref{fig:part} (c)), that is features were not extracted at different resolutions. Additionally, we also reduce the number of layers in each dense block so that we only have three blocks, each has $D=6$ number of layers and $k=24$ number of filters in each layer (growth rate). This is necessary because it has to be trained parallel to the GAN. Otherwise, an increase of the run time and a reduction of the largest possible batch size can be observed. The loss value described in equation \ref{eq:task_loss} is added to the generator loss $L_G$ and to the discriminator loss $L_D$, but the parameters of the task model are only updated during the optimization step of the discriminator.

\section{Datasets}\label{sec:data}

We use several datasets from different sources and therefore various complexities in terms of sequence length and number of different symbols.

\noindent \textbf{Im2Latex-100k} \cite{deng2016wygiswys} contains \num{100000} real world \latex equations extracted from publicly available pre-prints from the arxiv.org server. Compared to the other datasets considered, the set contains \num{380} different symbols and has an average formula length of \num{64} tokens. 

\noindent \textbf{CROHME 14} \cite{crohme2016} is a data collection with handwritten formulas and contains more than \num{10000} samples. The dataset was collected using pen-based devices, tablets or touch devices so that only the stroke information was stored and no pre-rendered images are available. The examples in this record have an average of \num{14} tokens and consist of \num{126} different symbols. 


\noindent \textbf{NTCIR-12 MathIR} \cite{zanibbi2016ntcir} (	NII (National Institute of Informatics) Test Collection for Information Resources) is a dataset for the retrieval of mathematical formulas and documents from arXiv.org or Wikipedia. The authors provide different dumps of both sources. After parsing all Wikipedia dump files, normalizing and filtering for formulas, \num{470009} formulas remain. This dataset contains the highest variation with \num{573} different symbols, but the formulas are very short with an average of \num{12} tokens.

As a pre-processing step, we have parsed all formulas using the Katex \cite{KaTeX} software project. Initially, we excluded all equations from our training set that could not be parsed, this case occurs mainly in the NTCIR-12 MathIR dataset. The excluded formulas contain a \latex command to change positions or text styles that are not supported by the Katex project. In the second step, we split the \latex formulas in tokens and remove all additional spacing commands such as \textbackslash, or \textbackslash quad. Finally, we use the Katex parser to render the formulas to a corresponding image. To prevent our model from overfitting we use different fonts, font sizes and spacings between the symbols. For the handwritten records we use the stroke information to create images for this domain.

For the Katex parser we randomly sample from  the following parameter space (padding and font size given in pixel):

\begin{description}
    \item \textbf{Font:} 'mathsf', 'mathtt', 'mathit', 'mathbf', 'mathrm', 'mathnormal' or 'textstyle'
    \item \textbf{Padding:}  \SI{0}{px}, \SI{1}{px}, \dots, \SI{15}{px}
    \item \textbf{Font size:} \SI{16}{px}, \SI{17}{px}, \dots, \SI{50}{px}
\end{description}

During the training process, additional data augmentation techniques are randomly applied to the datasets. The images are rotated in the range of $\pm 4^\circ$, sheared, randomly padded \num{10}-\SI{20}{px} at the boarder of the image, and randomly scaled by changing the aspect ratio. All samples are normalized in a way that the intensity of the strokes is close to one and the background intensity is close to zero, which reduces the input bias. 

These pre-processing steps are necessary because particularly the CROHME dataset contains only a small number of formulas and thus even fewer examples of the individual symbols are available. 

\section{Experimental Results}\label{sec:exp}
In the following, we investigate the individual components of our proposed system. Training details and qualitative results for generated formulas using the proposed GAN approach are presented in Section~\ref{sec:exp_gan}. A quantitative analysis is conducted in Section~\ref{sec:ablation} to evaluate the impact of the various GAN components. Ultimately, the results for formula recognition are reported and compared to the related work in Section~\ref{sec:recognition}.

\subsection{Synthesization of Handwritten Formulas}\label{sec:exp_gan}
\begin{figure*}
\centering
  \includegraphics[width=0.9\textwidth]{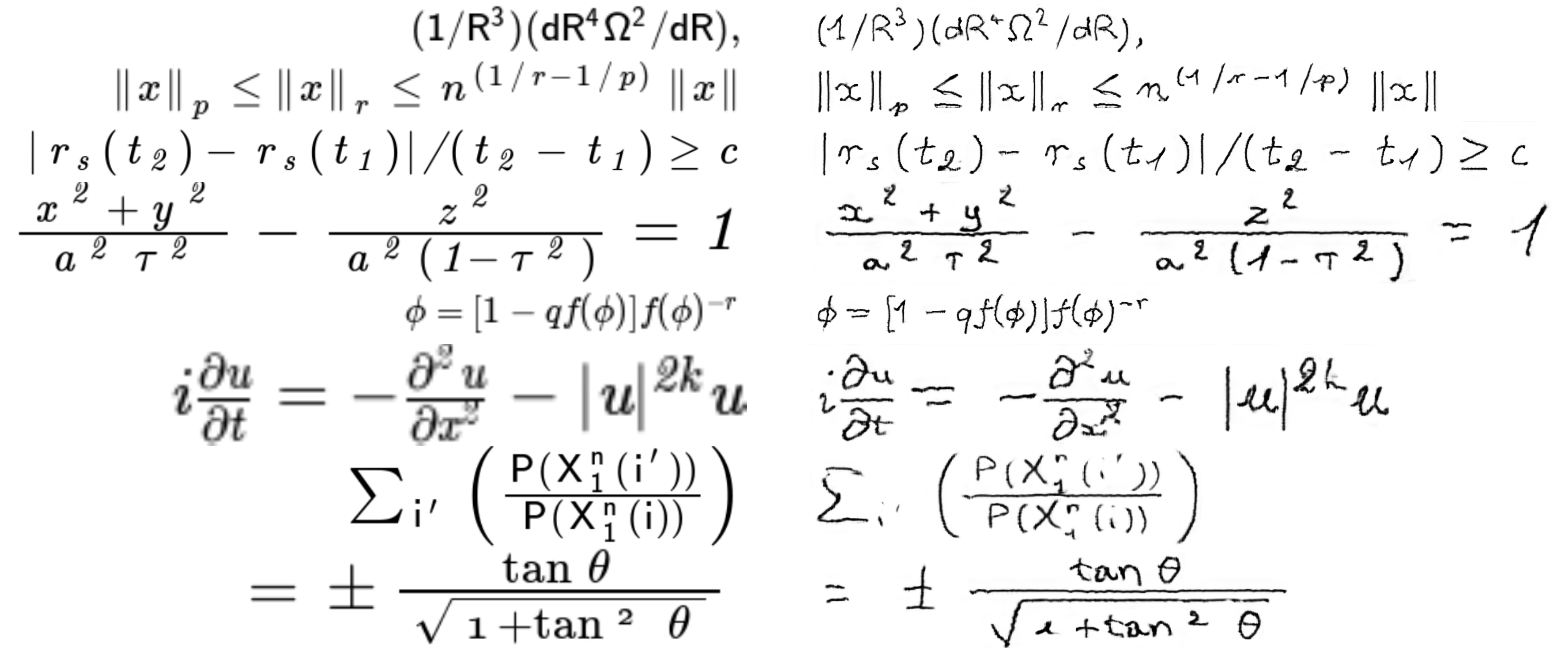}
  \caption{Example results generated by the proposed GAN model after \num{90000} iterations from the extracted NTCIR-12 MathIR dataset~\cite{zanibbi2016ntcir}. The formulas on the left side show the input pictures for the generator and the right side shows the corresponding synthesized handwritten image.}
  \label{fig:gan_result_ntcir}
\end{figure*}

In the first experiment, we evaluate the quality of the handwritten formulas synthesized by the GAN. The Katex rendered CROHME tex code,  handwritten CROHME equations and the rendered NTCIR-12 MathIR images are used as the input for the generator as well as samples for the discriminator for real images. Due to memory limitations, all input images are scaled to a height of \SI{128}{px}. Afterwards all formulas are sorted out, if the width exceeds \SI{512}{px} or their length is larger than \num{50} tokens. To solve the optimization problem, we use the ADAM (adaptive moment estimation) algorithm~\cite{kingma2014adam} and a learning rate for the discriminator of \num{2e-4} and for the generator of \num{5e-5}.

After the training process, the trained generator is used to generate handwritten versions of the datasets described in Section \ref{sec:data}. In contrast to the training process, formulas larger than \SI{512}{px} are now also processed. The only limitation is that the product of width and height must be smaller than \num{640000}, so that it still fits into GPU memory. Additionally, to get even more variation for the synthesized data, we sample different \latex font sizes and types for the synthesis. Finally, this method generates \num{294702} new handwritten formulas for Im2Latex and \num{469988} for the NTCIR dataset.

\begin{figure*}
\centering
  \includegraphics[width=0.9\textwidth]{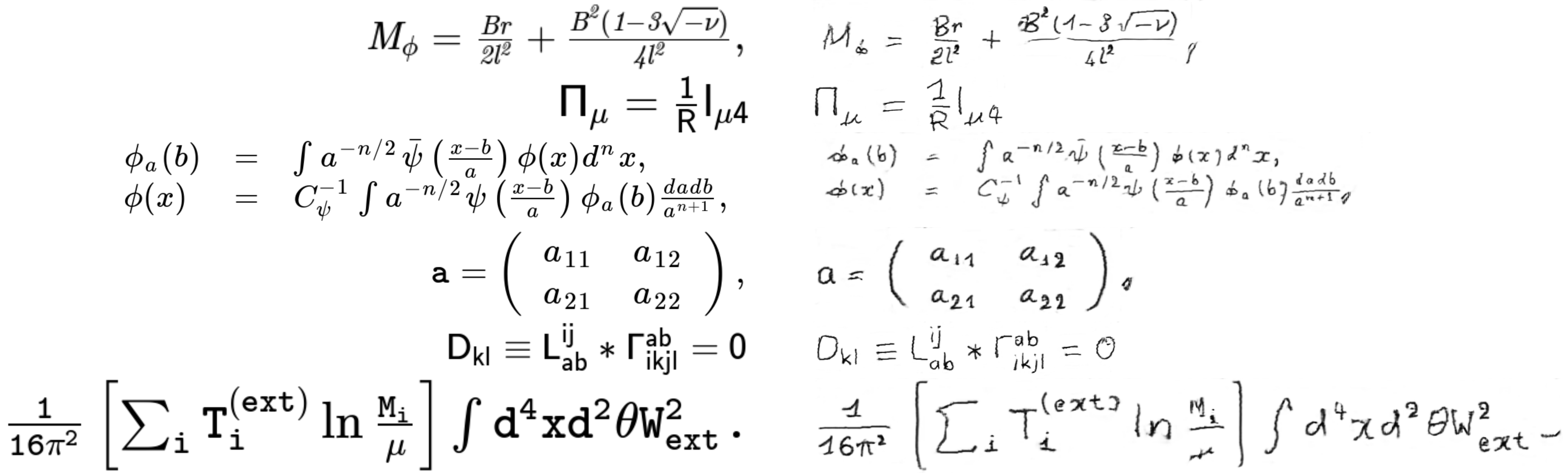}
  \caption{Sample images generated by the proposed GAN model after \num{90000} iterations from the extracted Im2Latex-100k dataset~\cite{deng2016wygiswys}. The formulas on the left side show the input pictures for the generator and the right side shows the corresponding synthesized handwritten image.}
  \label{fig:gan_result_im2latex}
\end{figure*}

The examples in Figure \ref{fig:gan_result_ntcir} and \ref{fig:gan_result_im2latex} illustrate that the proposed GAN model can transfer the input images to the other domain while retaining the content of the equation. 
The experimental results show that the neural network does not always generate the same representation for the same symbols, for example, the different equal signs in Figure \ref{fig:gan_result_ntcir}. Furthermore, the generator manages to create formula characters that it has never seen as a handwritten version before, like \textbackslash Gamma ($\Gamma$) or \textbackslash ast ($\ast$) in Figure \ref{fig:gan_result_im2latex}. But the results also shows some limitations, for example, some symbols become unreadable and smeared when they are very small or they stand very close to one another. Furthermore, the generator tends to create shaky strokes (i.e., waves instead of straight lines) and inserts some other artifacts. Some of these errors could be corrected by further post-processing or filtering steps.

\subsection{Evaluation of different network aspects}\label{sec:ablation}

In order to evaluate the contribution of different components and parameters to the performance of the GAN model, some experiments are performed. In each case a GAN is trained as described in Section \ref{sec:exp_gan}, and then a dataset is generated which only consists of the synthesized formulas of NTCIR-12 MathIR. Afterwards, the DenseNet classifier, which also serves as a task classifier in Section \ref{sec:exp_gan}, is trained on these formulas and these are subsequently evaluated on the CROHME dataset. The idea behind is that the GAN model which produces results most similar to the target data should achieve the best performance. At the same time it is evaluated if the generated symbols match the ground truth token sequence.

To evaluate the the models we use the perplexity metric, which corresponds to the exponential of the cross-entropy loss of the total sequence. A smaller perplexity value indicates a better result.

\begin{table}
    \centering
    \begin{tabular}{ll|rrr}
        &Architecture & \multicolumn{3}{c}{Perplexity} \\\hline
        &GAN iteration & \num{100000} & \num{200000} & \num{300000} \\\hline
        (1)& $\lambda = 0$ & \num{15.66} & \num{13.38} & \num{13.16}\\
        (2)& $\lambda = 1$ & \num{26.88} & \num{70.76} & \num{68.20}\\
        (3)& $\lambda = 2$ & \num{23.51} & \num{34.47} & \num{74.71}\\
        (4)& $\lambda = 10$ & \num{48.46} & \num{86.80} & \num{113.51}\\\hline
        (5)& no attention &          \num{77.20} & \num{20.59} & \num{81.12}\\\hline
        (6)& swap target domain &    \num{423.11} & \num{123.36} &  \num{24.30} \\\hline
        (7)& no handwritten source & \num{15.62}  & \num{7.34}   &  \num{9.16}  \\
    \end{tabular}
    \caption{Comparison of different architecture and training procedures for the formula GAN after a different number of iterations of GAN training. The models 5, 6 and 7 use $\lambda = 1$ during training.}
    
    \label{tab:comparision_architecture}
\end{table}

The results of the different architectures and parameter sets are shown in Table \ref{tab:comparision_architecture}. It is surprising that the model (1) without the task model ($\lambda = 0$) gives good results compared to other $\lambda$ values. 
This could be caused by several reasons. First, the GAN could overfit and produce a very accurate image, but of wrong but similar symbols. Taking into account that not all symbols are evenly distributed, this may cause the error to appear too small. Secondly, the task model used here might have disturbed the training of the GAN more than it helped. We will investigate this in further experiments in the following Section~\ref{sec:recognition}.

However, the best examined network (7) shows even better results. The main difference here is that the model at the input of the generator has never seen hand-written formulas.
The experiment without the attention module (5) shows that the performance could be increased in \num{2} of the \num{3} iterations when using the attention mechanism within the network. In setting (6) the target class of the generator was chosen randomly. As a consequence, it can happen that the source domain is the same as the target domain and the network would only have to restore the original image. The idea was to make it easy for the generator to train in the beginning, but the performance is much worse compared to the other models in the initial iterations.

Nevertheless, it should be noted that the perplexity metric does not directly show that each character is correct in the displayed sequence. Some characters could still be replaced by similar ones without affecting the perplexity metric greatly. 
Furthermore, it should be taken into account that from this experiment no statement can be derived on how closely the synthesised formulas appear to handwritten ones, but only how well the synthesised formulas can still be recognized.
For the training of the recognition model in the following section, we select the models (1), (2) and (7) after \num{300000} iterations each, since these models could achieve the highest performance in this first experiment.  

\subsection{Handwritten Formula Recognition}\label{sec:recognition}

\begin{table*}
    \centering
    \begin{tabular}{lll|cccc}
         &&& \multicolumn{2}{c}{normalize \SI{128}{px}} & \multicolumn{2}{c}{normalize symbol} \\\hline
        &Datasets & normalize&        WER & ExpRate  & WER & ExpRate\\\hline        
    (I)  &CROHME 14& 128px&                          \num{73.9} & \num{00.1} & \num{93.2} & \num{00.0}\\
    (II)  &CROHME 14\& GAN NTCIR-12 (1) & 128px&     \num{18.8} & \num{37.7} & \num{22.4} & \num{31.1}\\
    (III)  &CROHME 14\& GAN NTCIR-12 (2) & 128px&    \num{16.6} & \num{36.7} & \num{21.7} & \num{32.5}\\
    (IV)  &CROHME 14\& GAN NTCIR-12 (7) & 128px&     \num{16.4} & \num{40.5} & \num{22.0} & \num{32.8}\\\hline
    (V)&  CROHME 14& symbol&                         \num{36.1} & \num{11.9}  & \num{21.3} & \num{32.0}\\
    (VI)  &CROHME 14\& GAN NTCIR-12 (1) & symbol&    \num{30.8} & \num{29.2} & \num{22.5} & \num{32.3}\\
    (VII)  &CROHME 14\& GAN NTCIR-12 (2) & symbol&   \num{18.6} & \num{34.8} & \num{19.0} & \num{38.1}\\
    (VIII)  &CROHME 14\& GAN NTCIR-12 (7) & symbol&  \num{17.6} & \num{34.5} & \num{18.7} & \num{39.4}\\
    \end{tabular}
    \caption{Different results with the CROHME 14 testset. (I) and (V) Contains only samples from the original training dataset. (II-IV) and (VI-VIII) Additional synthesized samples based on the NTCIR-12 MathIR datasets are used to enlarge the dataset. The systems (V-VIII) are trained by normalized symbol height. Please note that only model (V) with normalized symbol was examined by Zhang et al.~\cite{zhang2018multi}}
    \label{tab:crohme_results}
\end{table*}

To evaluate the quality of the synthesized formulas, we analyze the usefulness of the newly generated data to train a state-of-the-art approach for handwritten formula recognition. To the best of our knowledge, the CROHME competition is the only dataset that provides a sufficient amount of training and test data for handwritten formula recognition to enable a fair evaluation. The DenseNet model of Zhang et al.~\cite{zhang2018multi} (task models used in the GAN in Section \ref{sec:task_model} with more layers and parameters) is trained and evaluated with different datasets. The baseline systems (I) and (V) use the training dataset provided by the CROHME 2014 recognition challenge. The other systems (II-IV) and (VI-VIII) sample equally distributed from the official CROHME training data and the synthesized handwritten images generated by the GAN for the NTCIR-12 MathIR competition. To avoid an impact on the performance on the benchmark, only the formulas are used which do not use any characters outside the CROHME dataset (\num{265871} images). The reason is that the CROHME dataset only contains \num{110} out of the \num{327} symbols used to train the generator. For example, the CROHME dataset does not contain the letter \emph{D} which could be incorrectly predicted by our model instead of the number \emph{0}.

For the experiments, we use a state-of-the-art approach for handwritten formula recognition, DenseNet implementation\footnote{Implementation from Wang: \url{https://github.com/whywhs/Pytorch-Handwritten-Mathematical-Expression-Recognition.git} \#9c5b248} of Zhang et al.~\cite{zhang2018multi}, and optimize the input for our datasets. All systems were optimized with a batch size of \num{6} and optimized with the SGD algorithm with momentum of \num{0.9}. The initial learning rate is \num{1e-4} and is reduced by a factor of \num{10} if there is no improvement on the validation dataset. The model with the highest Expression Recognition Rate (ExpRate) is then applied to the test set. Since the images synthesized by GAN are normalized to a height of \num{128} pixel, this yields a different height of the same character. Therefore the systems (I-IV) were trained with normalized input height of \num{128} pixel. In contrast, the formula images of systems (V-VIII) were scaled to the height of the rendered \latex image with a fixed font size, which should yield relatively equal symbol sizes.



During the test process we use a beam search decoder to predict the \latex sequence. The results show that the additional, synthetically generated training data has improved formula recognition in terms of Word Error Rate (WER) and Expression Recognition Rate (ExpRate). In order to carry out as many experiments as possible, we do not use the ensemble approach of Wang et al.~\cite{zhang2018multi} with a total of \num{5} models during the beam search inference. Instead, we use a single model for each beam search with a beam size of \num{10}. This explains the relatively large performance drop compared to the originally reported WER and ExpRate values. Another interesting result is that the system (I) trained with formula images of \SI{128}{px} height has not converged, indicating that for the CROHME dataset the size of the individual symbols is decisive for the performance, especially when there are comparatively few samples available. The results with the synthesized formulas show generally better results than those without synthesized training images. Moreover, the synthesized formulas are well suited to make models more robust for different resolutions of symbols. Furthermore, it is interesting to see that the model without the task model (1) with $\lambda=0$ does not perform better than the models with task recognition (2) with $\lambda=1$, in contrast to the ablation study in Section \ref{sec:ablation}.

\section{Conclusions}\label{sec:conclusion}

In this paper, we have presented a novel approach to generate a large number of synthesized handwritten formulas. The proposed approach is based on generative adversarial networks with self-attention and aims at enabling the large-scale training of deep learning systems for handwritten formula recognition. In our experiments we could show that it is possible to synthesize handwritten equations from a rendered \latex formula counterpart. Furthermore, the model is able to generate symbols for which it has never seen a training example in handwritten form. We have shown exemplary results from our synthesized dataset of handwritten formulas. Moreover, we have demonstrated that the incorporation of these training data can improve our implementation of a state-of-the-art approach for handwritten formula recognition on the CROHME 2014 benchmark dataset. We hope that the dataset provided will help close the gap between the recognition of handwritten and rendered equations. 

In the future, we plan to extend our training data with photos of formulas, so that we can also train our GAN in order to generate formulas with natural distortions and backgrounds. Such data will allow us to optimize the system for more complex scenarios such as formula recognition in lecture videos. Furthermore, it has been shown in practice that the availability of larger datasets makes it possible to use correspondingly more complex models.
\bibliographystyle{ACM-Reference-Format}
\bibliography{main}


\begin{thebibliography}{31}


\ifx \showCODEN    \undefined \def \showCODEN     #1{\unskip}     \fi
\ifx \showDOI      \undefined \def \showDOI       #1{#1}\fi
\ifx \showISBNx    \undefined \def \showISBNx     #1{\unskip}     \fi
\ifx \showISBNxiii \undefined \def \showISBNxiii  #1{\unskip}     \fi
\ifx \showISSN     \undefined \def \showISSN      #1{\unskip}     \fi
\ifx \showLCCN     \undefined \def \showLCCN      #1{\unskip}     \fi
\ifx \shownote     \undefined \def \shownote      #1{#1}          \fi
\ifx \showarticletitle \undefined \def \showarticletitle #1{#1}   \fi
\ifx \showURL      \undefined \def \showURL       {\relax}        \fi
\providecommand\bibfield[2]{#2}
\providecommand\bibinfo[2]{#2}
\providecommand\natexlab[1]{#1}
\providecommand\showeprint[2][]{arXiv:#2}

\bibitem[\protect\citeauthoryear{Academy}{Academy}{2017}]%
        {KaTeX}
\bibfield{author}{\bibinfo{person}{Khan Academy}.}
  \bibinfo{year}{2017}\natexlab{}.
\newblock \bibinfo{title}{KaTeX math typesetting library}.
\newblock
\newblock
\newblock
\shownote{\url{https://khan.github.io/KaTeX/}.}


\bibitem[\protect\citeauthoryear{Bahdanau, Cho, and Bengio}{Bahdanau
  et~al\mbox{.}}{2015}]%
        {bahdanau2014neural}
\bibfield{author}{\bibinfo{person}{Dzmitry Bahdanau},
  \bibinfo{person}{Kyunghyun Cho}, {and} \bibinfo{person}{Yoshua Bengio}.}
  \bibinfo{year}{2015}\natexlab{}.
\newblock \showarticletitle{Neural Machine Translation by Jointly Learning to
  Align and Translate}. In \bibinfo{booktitle}{\emph{3rd International
  Conference on Learning Representations, {ICLR} 2015, San Diego, CA, USA, May
  7-9, 2015, Conference Track Proceedings}},
  \bibfield{editor}{\bibinfo{person}{Yoshua Bengio} {and} \bibinfo{person}{Yann
  LeCun}} (Eds.).
\newblock
\urldef\tempurl%
\url{http://arxiv.org/abs/1409.0473}
\showURL{%
\tempurl}


\bibitem[\protect\citeauthoryear{Bousmalis, Silberman, Dohan, Erhan, and
  Krishnan}{Bousmalis et~al\mbox{.}}{2017}]%
        {bousmalis2016unsupervised}
\bibfield{author}{\bibinfo{person}{Konstantinos Bousmalis},
  \bibinfo{person}{Nathan Silberman}, \bibinfo{person}{David Dohan},
  \bibinfo{person}{Dumitru Erhan}, {and} \bibinfo{person}{Dilip Krishnan}.}
  \bibinfo{year}{2017}\natexlab{}.
\newblock \showarticletitle{Unsupervised Pixel-Level Domain Adaptation with
  Generative Adversarial Networks}. In \bibinfo{booktitle}{\emph{2017 {IEEE}
  Conference on Computer Vision and Pattern Recognition, {CVPR} 2017, Honolulu,
  HI, USA, July 21-26, 2017}}. \bibinfo{publisher}{{IEEE} Computer Society},
  \bibinfo{pages}{95--104}.
\newblock
\urldef\tempurl%
\url{https://doi.org/10.1109/CVPR.2017.18}
\showDOI{\tempurl}


\bibitem[\protect\citeauthoryear{Brock, Donahue, and Simonyan}{Brock
  et~al\mbox{.}}{2019}]%
        {brock2018large}
\bibfield{author}{\bibinfo{person}{Andrew Brock}, \bibinfo{person}{Jeff
  Donahue}, {and} \bibinfo{person}{Karen Simonyan}.}
  \bibinfo{year}{2019}\natexlab{}.
\newblock \showarticletitle{Large Scale {GAN} Training for High Fidelity
  Natural Image Synthesis}. In \bibinfo{booktitle}{\emph{7th International
  Conference on Learning Representations, {ICLR} 2019, New Orleans, LA, USA,
  May 6-9, 2019}}. \bibinfo{publisher}{OpenReview.net}.
\newblock
\urldef\tempurl%
\url{https://openreview.net/forum?id=B1xsqj09Fm}
\showURL{%
\tempurl}


\bibitem[\protect\citeauthoryear{Carter, Ha, Johnson, and Olah}{Carter
  et~al\mbox{.}}{2016}]%
        {carter2016experiments}
\bibfield{author}{\bibinfo{person}{Shan Carter}, \bibinfo{person}{David Ha},
  \bibinfo{person}{Ian Johnson}, {and} \bibinfo{person}{Chris Olah}.}
  \bibinfo{year}{2016}\natexlab{}.
\newblock \showarticletitle{Experiments in Handwriting with a Neural Network}.
\newblock \bibinfo{journal}{\emph{Distill}} (\bibinfo{year}{2016}).
\newblock
\urldef\tempurl%
\url{https://doi.org/10.23915/distill.00004}
\showDOI{\tempurl}


\bibitem[\protect\citeauthoryear{Davila and Zanibbi}{Davila and
  Zanibbi}{2018}]%
        {davila2018visual}
\bibfield{author}{\bibinfo{person}{Kenny Davila} {and} \bibinfo{person}{Richard
  Zanibbi}.} \bibinfo{year}{2018}\natexlab{}.
\newblock \showarticletitle{Visual Search Engine for Handwritten and Typeset
  Math in Lecture Videos and {LATEX} Notes}. In \bibinfo{booktitle}{\emph{16th
  International Conference on Frontiers in Handwriting Recognition, {ICFHR}
  2018, Niagara Falls, NY, USA, August 5-8, 2018}}. \bibinfo{publisher}{{IEEE}
  Computer Society}, \bibinfo{pages}{50--55}.
\newblock
\urldef\tempurl%
\url{https://doi.org/10.1109/ICFHR-2018.2018.00018}
\showDOI{\tempurl}


\bibitem[\protect\citeauthoryear{Deng, Kanervisto, Ling, and Rush}{Deng
  et~al\mbox{.}}{2017a}]%
        {deng2017image}
\bibfield{author}{\bibinfo{person}{Yuntian Deng}, \bibinfo{person}{Anssi
  Kanervisto}, \bibinfo{person}{Jeffrey Ling}, {and}
  \bibinfo{person}{Alexander~M. Rush}.} \bibinfo{year}{2017}\natexlab{a}.
\newblock \showarticletitle{Image-to-Markup Generation with Coarse-to-Fine
  Attention}. In \bibinfo{booktitle}{\emph{Proceedings of the 34th
  International Conference on Machine Learning, {ICML} 2017, Sydney, NSW,
  Australia, 6-11 August 2017}} \emph{(\bibinfo{series}{Proceedings of Machine
  Learning Research}, Vol.~\bibinfo{volume}{70})},
  \bibfield{editor}{\bibinfo{person}{Doina Precup} {and}
  \bibinfo{person}{Yee~Whye Teh}} (Eds.). \bibinfo{publisher}{{PMLR}},
  \bibinfo{pages}{980--989}.
\newblock
\urldef\tempurl%
\url{http://proceedings.mlr.press/v70/deng17a.html}
\showURL{%
\tempurl}


\bibitem[\protect\citeauthoryear{Deng, Kanervisto, and Rush}{Deng
  et~al\mbox{.}}{2016}]%
        {deng2016wygiswys}
\bibfield{author}{\bibinfo{person}{Yuntian Deng}, \bibinfo{person}{Anssi
  Kanervisto}, {and} \bibinfo{person}{Alexander~M. Rush}.}
  \bibinfo{year}{2016}\natexlab{}.
\newblock \showarticletitle{What You Get Is What You See: A Visual Markup
  Decompiler}.
\newblock \bibinfo{journal}{\emph{CoRR}} (\bibinfo{year}{2016}).
\newblock
\showeprint[arxiv]{1609.04938v1}
\urldef\tempurl%
\url{http://arxiv.org/abs/1609.04938v1}
\showURL{%
\tempurl}


\bibitem[\protect\citeauthoryear{Deng, Yu, Yao, and Sun}{Deng
  et~al\mbox{.}}{2017b}]%
        {deng2017attention}
\bibfield{author}{\bibinfo{person}{Yukai Deng}, \bibinfo{person}{Yao Yu},
  \bibinfo{person}{Junfeng Yao}, {and} \bibinfo{person}{Changyin Sun}.}
  \bibinfo{year}{2017}\natexlab{b}.
\newblock \showarticletitle{An attention based image to latex markup decoder}.
  In \bibinfo{booktitle}{\emph{2017 Chinese Automation Congress (CAC)}}. IEEE,
  \bibinfo{pages}{7199--7203}.
\newblock


\bibitem[\protect\citeauthoryear{Goodfellow, Pouget{-}Abadie, Mirza, Xu,
  Warde{-}Farley, Ozair, Courville, and Bengio}{Goodfellow
  et~al\mbox{.}}{2014}]%
        {goodfellow2014generative}
\bibfield{author}{\bibinfo{person}{Ian~J. Goodfellow}, \bibinfo{person}{Jean
  Pouget{-}Abadie}, \bibinfo{person}{Mehdi Mirza}, \bibinfo{person}{Bing Xu},
  \bibinfo{person}{David Warde{-}Farley}, \bibinfo{person}{Sherjil Ozair},
  \bibinfo{person}{Aaron~C. Courville}, {and} \bibinfo{person}{Yoshua Bengio}.}
  \bibinfo{year}{2014}\natexlab{}.
\newblock \showarticletitle{Generative Adversarial Nets}. In
  \bibinfo{booktitle}{\emph{Advances in Neural Information Processing Systems
  27: Annual Conference on Neural Information Processing Systems 2014, December
  8-13 2014, Montreal, Quebec, Canada}},
  \bibfield{editor}{\bibinfo{person}{Zoubin Ghahramani}, \bibinfo{person}{Max
  Welling}, \bibinfo{person}{Corinna Cortes}, \bibinfo{person}{Neil~D.
  Lawrence}, {and} \bibinfo{person}{Kilian~Q. Weinberger}} (Eds.).
  \bibinfo{pages}{2672--2680}.
\newblock
\urldef\tempurl%
\url{https://proceedings.neurips.cc/paper/2014/hash/5ca3e9b122f61f8f06494c97b1afccf3-Abstract.html}
\showURL{%
\tempurl}


\bibitem[\protect\citeauthoryear{Graves}{Graves}{2013}]%
        {graves2013generating}
\bibfield{author}{\bibinfo{person}{Alex Graves}.}
  \bibinfo{year}{2013}\natexlab{}.
\newblock \showarticletitle{Generating Sequences With Recurrent Neural
  Networks}.
\newblock \bibinfo{journal}{\emph{CoRR}}  \bibinfo{volume}{abs/1308.0850}
  (\bibinfo{year}{2013}).
\newblock
\showeprint[arxiv]{1308.0850}
\urldef\tempurl%
\url{http://arxiv.org/abs/1308.0850}
\showURL{%
\tempurl}


\bibitem[\protect\citeauthoryear{He, Zhang, Ren, and Sun}{He
  et~al\mbox{.}}{2016}]%
        {he2016deep}
\bibfield{author}{\bibinfo{person}{Kaiming He}, \bibinfo{person}{Xiangyu
  Zhang}, \bibinfo{person}{Shaoqing Ren}, {and} \bibinfo{person}{Jian Sun}.}
  \bibinfo{year}{2016}\natexlab{}.
\newblock \showarticletitle{Deep Residual Learning for Image Recognition}. In
  \bibinfo{booktitle}{\emph{2016 {IEEE} Conference on Computer Vision and
  Pattern Recognition, {CVPR} 2016, Las Vegas, NV, USA, June 27-30, 2016}}.
  \bibinfo{publisher}{{IEEE} Computer Society}, \bibinfo{pages}{770--778}.
\newblock
\urldef\tempurl%
\url{https://doi.org/10.1109/CVPR.2016.90}
\showDOI{\tempurl}


\bibitem[\protect\citeauthoryear{Hochreiter and Schmidhuber}{Hochreiter and
  Schmidhuber}{1997}]%
        {hochreiter1997long}
\bibfield{author}{\bibinfo{person}{Sepp Hochreiter} {and}
  \bibinfo{person}{J{\"{u}}rgen Schmidhuber}.} \bibinfo{year}{1997}\natexlab{}.
\newblock \showarticletitle{Long Short-Term Memory}.
\newblock \bibinfo{journal}{\emph{Neural Comput.}} \bibinfo{volume}{9},
  \bibinfo{number}{8} (\bibinfo{year}{1997}), \bibinfo{pages}{1735--1780}.
\newblock
\urldef\tempurl%
\url{https://doi.org/10.1162/neco.1997.9.8.1735}
\showDOI{\tempurl}


\bibitem[\protect\citeauthoryear{Huang, Liu, van~der Maaten, and
  Weinberger}{Huang et~al\mbox{.}}{2017}]%
        {huang2017densely}
\bibfield{author}{\bibinfo{person}{Gao Huang}, \bibinfo{person}{Zhuang Liu},
  \bibinfo{person}{Laurens van~der Maaten}, {and} \bibinfo{person}{Kilian~Q.
  Weinberger}.} \bibinfo{year}{2017}\natexlab{}.
\newblock \showarticletitle{Densely Connected Convolutional Networks}.
\newblock  (\bibinfo{year}{2017}), \bibinfo{pages}{2261--2269}.
\newblock
\urldef\tempurl%
\url{https://doi.org/10.1109/CVPR.2017.243}
\showDOI{\tempurl}


\bibitem[\protect\citeauthoryear{Kingma and Ba}{Kingma and Ba}{2015}]%
        {kingma2014adam}
\bibfield{author}{\bibinfo{person}{Diederik~P. Kingma} {and}
  \bibinfo{person}{Jimmy Ba}.} \bibinfo{year}{2015}\natexlab{}.
\newblock \showarticletitle{Adam: {A} Method for Stochastic Optimization}. In
  \bibinfo{booktitle}{\emph{3rd International Conference on Learning
  Representations, {ICLR} 2015, San Diego, CA, USA, May 7-9, 2015, Conference
  Track Proceedings}}, \bibfield{editor}{\bibinfo{person}{Yoshua Bengio} {and}
  \bibinfo{person}{Yann LeCun}} (Eds.).
\newblock
\urldef\tempurl%
\url{http://arxiv.org/abs/1412.6980}
\showURL{%
\tempurl}


\bibitem[\protect\citeauthoryear{Kumar, Kandala, and Reddy}{Kumar
  et~al\mbox{.}}{2018}]%
        {kumar2018synthesizing}
\bibfield{author}{\bibinfo{person}{K.~Manoj Kumar}, \bibinfo{person}{Harish
  Kandala}, {and} \bibinfo{person}{N.~Sudhakar Reddy}.}
  \bibinfo{year}{2018}\natexlab{}.
\newblock \showarticletitle{Synthesizing and Imitating Handwriting Using Deep
  Recurrent Neural Networks and Mixture Density Networks}. In
  \bibinfo{booktitle}{\emph{9th International Conference on Computing,
  Communication and Networking Technologies, {ICCCNT} 2018, Bengaluru, India,
  July 10-12, 2018}}. \bibinfo{publisher}{{IEEE}}, \bibinfo{pages}{1--6}.
\newblock
\urldef\tempurl%
\url{https://doi.org/10.1109/ICCCNT.2018.8493843}
\showDOI{\tempurl}


\bibitem[\protect\citeauthoryear{Le, Indurkhya, and Nakagawa}{Le
  et~al\mbox{.}}{2019}]%
        {le2019pattern}
\bibfield{author}{\bibinfo{person}{Anh~Duc Le}, \bibinfo{person}{Bipin
  Indurkhya}, {and} \bibinfo{person}{Masaki Nakagawa}.}
  \bibinfo{year}{2019}\natexlab{}.
\newblock \showarticletitle{Pattern generation strategies for improving
  recognition of Handwritten Mathematical Expressions}.
\newblock \bibinfo{journal}{\emph{Pattern Recognit. Lett.}}
  \bibinfo{volume}{128} (\bibinfo{year}{2019}), \bibinfo{pages}{255--262}.
\newblock
\urldef\tempurl%
\url{https://doi.org/10.1016/j.patrec.2019.09.002}
\showDOI{\tempurl}


\bibitem[\protect\citeauthoryear{Liu, Breuel, and Kautz}{Liu
  et~al\mbox{.}}{2017}]%
        {liu2017unsupervised}
\bibfield{author}{\bibinfo{person}{Ming{-}Yu Liu}, \bibinfo{person}{Thomas
  Breuel}, {and} \bibinfo{person}{Jan Kautz}.} \bibinfo{year}{2017}\natexlab{}.
\newblock \showarticletitle{Unsupervised Image-to-Image Translation Networks}.
  In \bibinfo{booktitle}{\emph{Advances in Neural Information Processing
  Systems 30: Annual Conference on Neural Information Processing Systems 2017,
  December 4-9, 2017, Long Beach, CA, {USA}}},
  \bibfield{editor}{\bibinfo{person}{Isabelle Guyon}, \bibinfo{person}{Ulrike
  von Luxburg}, \bibinfo{person}{Samy Bengio}, \bibinfo{person}{Hanna~M.
  Wallach}, \bibinfo{person}{Rob Fergus}, \bibinfo{person}{S.~V.~N.
  Vishwanathan}, {and} \bibinfo{person}{Roman Garnett}} (Eds.).
  \bibinfo{pages}{700--708}.
\newblock
\urldef\tempurl%
\url{https://proceedings.neurips.cc/paper/2017/hash/dc6a6489640ca02b0d42dabeb8e46bb7-Abstract.html}
\showURL{%
\tempurl}


\bibitem[\protect\citeauthoryear{Mirza and Osindero}{Mirza and
  Osindero}{2014}]%
        {mirza2014conditional}
\bibfield{author}{\bibinfo{person}{Mehdi Mirza} {and} \bibinfo{person}{Simon
  Osindero}.} \bibinfo{year}{2014}\natexlab{}.
\newblock \showarticletitle{Conditional Generative Adversarial Nets}.
\newblock \bibinfo{journal}{\emph{CoRR}}  \bibinfo{volume}{abs/1411.1784}
  (\bibinfo{year}{2014}).
\newblock
\showeprint[arxiv]{1411.1784}
\urldef\tempurl%
\url{http://arxiv.org/abs/1411.1784}
\showURL{%
\tempurl}


\bibitem[\protect\citeauthoryear{Miyato, Kataoka, Koyama, and Yoshida}{Miyato
  et~al\mbox{.}}{2018}]%
        {miyato2018spectral}
\bibfield{author}{\bibinfo{person}{Takeru Miyato}, \bibinfo{person}{Toshiki
  Kataoka}, \bibinfo{person}{Masanori Koyama}, {and} \bibinfo{person}{Yuichi
  Yoshida}.} \bibinfo{year}{2018}\natexlab{}.
\newblock \showarticletitle{Spectral Normalization for Generative Adversarial
  Networks}. In \bibinfo{booktitle}{\emph{6th International Conference on
  Learning Representations, {ICLR} 2018, Vancouver, BC, Canada, April 30 - May
  3, 2018, Conference Track Proceedings}}. \bibinfo{publisher}{OpenReview.net}.
\newblock
\urldef\tempurl%
\url{https://openreview.net/forum?id=B1QRgziT-}
\showURL{%
\tempurl}


\bibitem[\protect\citeauthoryear{Miyato and Koyama}{Miyato and Koyama}{2018}]%
        {miyato2018cgans}
\bibfield{author}{\bibinfo{person}{Takeru Miyato} {and}
  \bibinfo{person}{Masanori Koyama}.} \bibinfo{year}{2018}\natexlab{}.
\newblock \showarticletitle{cGANs with Projection Discriminator}. In
  \bibinfo{booktitle}{\emph{6th International Conference on Learning
  Representations, {ICLR} 2018, Vancouver, BC, Canada, April 30 - May 3, 2018,
  Conference Track Proceedings}}. \bibinfo{publisher}{OpenReview.net}.
\newblock
\urldef\tempurl%
\url{https://openreview.net/forum?id=ByS1VpgRZ}
\showURL{%
\tempurl}


\bibitem[\protect\citeauthoryear{Mouch{\`{e}}re, Viard{-}Gaudin, Zanibbi, and
  Garain}{Mouch{\`{e}}re et~al\mbox{.}}{2016}]%
        {crohme2016}
\bibfield{author}{\bibinfo{person}{Harold Mouch{\`{e}}re},
  \bibinfo{person}{Christian Viard{-}Gaudin}, \bibinfo{person}{Richard
  Zanibbi}, {and} \bibinfo{person}{Utpal Garain}.}
  \bibinfo{year}{2016}\natexlab{}.
\newblock \showarticletitle{{ICFHR2016} {CROHME:} Competition on Recognition of
  Online Handwritten Mathematical Expressions}. In
  \bibinfo{booktitle}{\emph{15th International Conference on Frontiers in
  Handwriting Recognition, {ICFHR} 2016, Shenzhen, China, October 23-26,
  2016}}. \bibinfo{publisher}{{IEEE} Computer Society},
  \bibinfo{pages}{607--612}.
\newblock
\urldef\tempurl%
\url{https://doi.org/10.1109/ICFHR.2016.0116}
\showDOI{\tempurl}


\bibitem[\protect\citeauthoryear{Russakovsky, Deng, Su, Krause, Satheesh, Ma,
  Huang, Karpathy, Khosla, Bernstein, Berg, and Li}{Russakovsky
  et~al\mbox{.}}{2015}]%
        {ILSVRC15}
\bibfield{author}{\bibinfo{person}{Olga Russakovsky}, \bibinfo{person}{Jia
  Deng}, \bibinfo{person}{Hao Su}, \bibinfo{person}{Jonathan Krause},
  \bibinfo{person}{Sanjeev Satheesh}, \bibinfo{person}{Sean Ma},
  \bibinfo{person}{Zhiheng Huang}, \bibinfo{person}{Andrej Karpathy},
  \bibinfo{person}{Aditya Khosla}, \bibinfo{person}{Michael~S. Bernstein},
  \bibinfo{person}{Alexander~C. Berg}, {and} \bibinfo{person}{Fei{-}Fei Li}.}
  \bibinfo{year}{2015}\natexlab{}.
\newblock \showarticletitle{ImageNet Large Scale Visual Recognition Challenge}.
\newblock \bibinfo{journal}{\emph{Int. J. Comput. Vis.}} \bibinfo{volume}{115},
  \bibinfo{number}{3} (\bibinfo{year}{2015}), \bibinfo{pages}{211--252}.
\newblock
\urldef\tempurl%
\url{https://doi.org/10.1007/s11263-015-0816-y}
\showDOI{\tempurl}


\bibitem[\protect\citeauthoryear{Sun, Shrivastava, Singh, and Gupta}{Sun
  et~al\mbox{.}}{2017}]%
        {sun2017revisiting}
\bibfield{author}{\bibinfo{person}{Chen Sun}, \bibinfo{person}{Abhinav
  Shrivastava}, \bibinfo{person}{Saurabh Singh}, {and} \bibinfo{person}{Abhinav
  Gupta}.} \bibinfo{year}{2017}\natexlab{}.
\newblock \showarticletitle{Revisiting Unreasonable Effectiveness of Data in
  Deep Learning Era}. In \bibinfo{booktitle}{\emph{{IEEE} International
  Conference on Computer Vision, {ICCV} 2017, Venice, Italy, October 22-29,
  2017}}. \bibinfo{publisher}{{IEEE} Computer Society},
  \bibinfo{pages}{843--852}.
\newblock
\urldef\tempurl%
\url{https://doi.org/10.1109/ICCV.2017.97}
\showDOI{\tempurl}


\bibitem[\protect\citeauthoryear{Wang and Liu}{Wang and Liu}{2019}]%
        {DBLP:journals/corr/abs-1908-11415}
\bibfield{author}{\bibinfo{person}{Zelun Wang} {and}
  \bibinfo{person}{Jyh{-}Charn Liu}.} \bibinfo{year}{2019}\natexlab{}.
\newblock \showarticletitle{Translating Mathematical Formula Images to LaTeX
  Sequences Using Deep Neural Networks with Sequence-level Training}.
\newblock \bibinfo{journal}{\emph{CoRR}}  \bibinfo{volume}{abs/1908.11415}
  (\bibinfo{year}{2019}).
\newblock
\showeprint[arxiv]{1908.11415}
\urldef\tempurl%
\url{http://arxiv.org/abs/1908.11415}
\showURL{%
\tempurl}


\bibitem[\protect\citeauthoryear{Wu, Yin, Zhang, Zhang, and Liu}{Wu
  et~al\mbox{.}}{2020}]%
        {wu2020handwritten}
\bibfield{author}{\bibinfo{person}{Jin-Wen Wu}, \bibinfo{person}{Fei Yin},
  \bibinfo{person}{Yan-Ming Zhang}, \bibinfo{person}{Xu-Yao Zhang}, {and}
  \bibinfo{person}{Cheng-Lin Liu}.} \bibinfo{year}{2020}\natexlab{}.
\newblock \showarticletitle{Handwritten Mathematical Expression Recognition via
  Paired Adversarial Learning}.
\newblock \bibinfo{journal}{\emph{International Journal of Computer Vision}}
  (\bibinfo{year}{2020}), \bibinfo{pages}{1--16}.
\newblock


\bibitem[\protect\citeauthoryear{Zanibbi, Aizawa, Kohlhase, Ounis, Topic, and
  Davila}{Zanibbi et~al\mbox{.}}{2016}]%
        {zanibbi2016ntcir}
\bibfield{author}{\bibinfo{person}{Richard Zanibbi}, \bibinfo{person}{Akiko
  Aizawa}, \bibinfo{person}{Michael Kohlhase}, \bibinfo{person}{Iadh Ounis},
  \bibinfo{person}{Goran Topic}, {and} \bibinfo{person}{Kenny Davila}.}
  \bibinfo{year}{2016}\natexlab{}.
\newblock \showarticletitle{{NTCIR-12} MathIR Task Overview}. In
  \bibinfo{booktitle}{\emph{Proceedings of the 12th {NTCIR} Conference on
  Evaluation of Information Access Technologies, National Center of Sciences,
  Tokyo, Japan, June 7-10, 2016}}, \bibfield{editor}{\bibinfo{person}{Noriko
  Kando}, \bibinfo{person}{Tetsuya Sakai}, {and} \bibinfo{person}{Mark
  Sanderson}} (Eds.). \bibinfo{publisher}{National Institute of Informatics
  {(NII)}}.
\newblock
\urldef\tempurl%
\url{http://research.nii.ac.jp/ntcir/workshop/OnlineProceedings12/pdf/ntcir/OVERVIEW/01-NTCIR12-OV-MathIR-ZanibbiR.pdf}
\showURL{%
\tempurl}


\bibitem[\protect\citeauthoryear{Zhang, Goodfellow, Metaxas, and Odena}{Zhang
  et~al\mbox{.}}{2018b}]%
        {zhang2018self}
\bibfield{author}{\bibinfo{person}{Han Zhang}, \bibinfo{person}{Ian~J.
  Goodfellow}, \bibinfo{person}{Dimitris~N. Metaxas}, {and}
  \bibinfo{person}{Augustus Odena}.} \bibinfo{year}{2018}\natexlab{b}.
\newblock \showarticletitle{Self-Attention Generative Adversarial Networks}.
\newblock \bibinfo{journal}{\emph{CoRR}}  \bibinfo{volume}{abs/1805.08318}
  (\bibinfo{year}{2018}).
\newblock
\showeprint[arxiv]{1805.08318}
\urldef\tempurl%
\url{http://arxiv.org/abs/1805.08318}
\showURL{%
\tempurl}


\bibitem[\protect\citeauthoryear{Zhang, Du, and Dai}{Zhang
  et~al\mbox{.}}{2018a}]%
        {zhang2018multi}
\bibfield{author}{\bibinfo{person}{Jianshu Zhang}, \bibinfo{person}{Jun Du},
  {and} \bibinfo{person}{Lirong Dai}.} \bibinfo{year}{2018}\natexlab{a}.
\newblock \showarticletitle{Multi-Scale Attention with Dense Encoder for
  Handwritten Mathematical Expression Recognition}. In
  \bibinfo{booktitle}{\emph{24th International Conference on Pattern
  Recognition, {ICPR} 2018, Beijing, China, August 20-24, 2018}}.
  \bibinfo{publisher}{{IEEE} Computer Society}, \bibinfo{pages}{2245--2250}.
\newblock
\urldef\tempurl%
\url{https://doi.org/10.1109/ICPR.2018.8546031}
\showDOI{\tempurl}


\bibitem[\protect\citeauthoryear{Zhang, Du, Zhang, Liu, Hu, Hu, Wei, and
  Dai}{Zhang et~al\mbox{.}}{2017}]%
        {zhang2017watch}
\bibfield{author}{\bibinfo{person}{Jianshu Zhang}, \bibinfo{person}{Jun Du},
  \bibinfo{person}{Shiliang Zhang}, \bibinfo{person}{Dan Liu},
  \bibinfo{person}{Yulong Hu}, \bibinfo{person}{Jin{-}Shui Hu},
  \bibinfo{person}{Si Wei}, {and} \bibinfo{person}{Li{-}Rong Dai}.}
  \bibinfo{year}{2017}\natexlab{}.
\newblock \showarticletitle{Watch, attend and parse: An end-to-end neural
  network based approach to handwritten mathematical expression recognition}.
\newblock \bibinfo{journal}{\emph{Pattern Recognit.}}  \bibinfo{volume}{71}
  (\bibinfo{year}{2017}), \bibinfo{pages}{196--206}.
\newblock
\urldef\tempurl%
\url{https://doi.org/10.1016/j.patcog.2017.06.017}
\showDOI{\tempurl}


\bibitem[\protect\citeauthoryear{Zhang, Bai, and Zhu}{Zhang
  et~al\mbox{.}}{2019}]%
        {zhang2019improved}
\bibfield{author}{\bibinfo{person}{Wei Zhang}, \bibinfo{person}{Zhiqiang Bai},
  {and} \bibinfo{person}{Yuesheng Zhu}.} \bibinfo{year}{2019}\natexlab{}.
\newblock \showarticletitle{An improved approach based on CNN-RNNs for
  mathematical expression recognition}. In
  \bibinfo{booktitle}{\emph{Proceedings of the 2019 4th International
  Conference on Multimedia Systems and Signal Processing}}.
  \bibinfo{pages}{57--61}.
\newblock


\end{thebibliography}

\appendix

\end{document}